%% file: iclr2025_conference.tex
\documentclass{article} 
\usepackage{iclr2025_conference,times}
\setcitestyle{numbers,square}

\input{math_commands.tex}

\usepackage[T1]{fontenc}
\usepackage[utf8]{inputenc}

\usepackage{hyperref}
\usepackage{url}
\usepackage{booktabs}
\usepackage{tabularx}
\usepackage{graphicx}
\usepackage{caption}
\usepackage{amssymb}
\usepackage{enumitem}
\usepackage[font=small,labelfont=bf]{caption}

\title{Anatomy of a Failure: When, How, and Why Deep Vision Fails in Scientific Domains}

\author{Ji-Hun Oh\textsuperscript{1}, Dou Hoon Kwark\textsuperscript{2}, Kianoush Falahkheirkhah\textsuperscript{3}, Kevin Yeh\textsuperscript{3},\\ \textbf{John Cheville\textsuperscript{4}, Volodymyr Kindratenko\textsuperscript{2,5,6}, Rohit Bhargava\textsuperscript{1,3,5,7,8,9,10,11}} \\ \\
    \small{\textsuperscript{1}Department of Mechanical Science and Engineering, University of Illinois Urbana-Champaign, IL, US}\\
    \small{\textsuperscript{2}Siebel School of Computing and Data Science, University of Illinois Urbana-Champaign, IL, US}\\
  \small{\textsuperscript{3}Beckman Institute for Advanced Science and Technology, University of Illinois Urbana-Champaign, IL, US}\\
  \small{\textsuperscript{4}Mayo Clinic, Rochester, MN, US}\\
  \small{\textsuperscript{5}Department of Electrical and Computer Engineering, University of Illinois Urbana-Champaign, IL, US}\\
  \small{\textsuperscript{6}National Center for Supercomputing Applications, University of Illinois Urbana-Champaign, IL, US}\\
  \small{\textsuperscript{7}Department of Bioengineering, University of Illinois Urbana-Champaign, IL, US}\\
  \small{\textsuperscript{8}Department of Chemical and Biomolecular Engineering, University of Illinois Urbana-Champaign, IL, US}\\
  \small{\textsuperscript{9}Department of Chemistry, University of Illinois Urbana-Champaign, IL, US}\\
 \small{\textsuperscript{10}Cancer Center at Illinois, University of Illinois Urbana-Champaign, IL, US}\\
\small{\textsuperscript{11}CZ Biohub Chicago, LLC, Chicago, IL, US}\\
}

\iclrfinalcopy 
\begin{document}

\maketitle

\begin{abstract}
Mirroring its ubiquity in popular media and all human activities, the use of deep learning (DL) is rapidly growing in scientific imaging modalities. However, unlike everyday RGB pictures, pixels encode precise physicochemical properties in scientific imaging across potentially thousands of channels. While DL is well validated on human-centric RGB perceptual tasks, its effectiveness for scientific imaging remains uncertain. Here, we show that the naive application of DL frameworks to scientific images can lead to critical failures. We evaluate the use of DL for pathology, comparing RGB images of stained tissue with the quantitative and information-rich biochemical signatures of infrared (IR) imaging. Despite this informational advantage, DL models trained on IR data paradoxically underperform. We investigate this discrepancy to find that IR data priors interact poorly with the simplicity bias of DL, causing models to collapse to one-dimensional predictions. This constitutes a catastrophic DL failure because the model’s representational capacity remains largely unused, while furthermore raising AI safety concerns and undermining the advantages of such scientific modalities. Notably, this problem persists even with state-of-the-art DL robustification strategies, which are primarily designed and validated for RGB imagery and thus inherit the same prior–bias mismatch. This work establishes a framework for understanding the limitations of generic DL in science and advocates for the study of modality-specific failure modes to guide the development of specialized, safe AI algorithms.
\end{abstract}

\section{Introduction}
Scientific and engineering progress increasingly depends on computational sensing modalities that extend beyond the limits of human RGB perception. Spanning the electromagnetic spectrum—from X-ray to infrared and terahertz radiation—as well as non-optical domains such as acoustic impedance and thermodynamic properties, scientific imaging modalities map physicochemical attributes onto digital arrays through specialized contrast mechanisms. The result is a complex yet information-rich spectral-spatial image. Because such signals are non-canonical and often unintuitive to human observers, machine vision becomes essential for translating raw measurements into interpretable knowledge and enabling the high-throughput analysis required in real-time applications such as diagnostic screening.

Deep Learning (DL), particularly through Convolutional Neural Networks (CNNs), has become the de facto framework for vision tasks. Accordingly, these methods are increasingly adopted for specialized sensing modalities, motivated by the promise of combining rich imaging content with the powerful pattern recognition capacity of DL. In practice, however, converting this theoretical synergy into practical performance gains remains challenging. A central issue is underspecification \cite{d2022underspecification}: DL models trained via Empirical Risk Minimization (ERM) converge to solutions that fit the training data, yet due to overparameterization, many disparate hypotheses can satisfy the ERM objective. In this regime, Inductive Biases (IBs)—algorithmic constraints shaping optimization trajectories and solution spaces—play a critical role in guiding learning toward generalizable solutions, ideally approximating the Bayes-optimal model. These biases include hard architectural priors (e.g., spatial locality and translation equivariance in CNNs) as well as soft priors such as data augmentation and regularization. However, the ERM+IB combination often remains insufficient, particularly in scientific domains characterized by scarce data and high-dimensional structure, where underspecification is exacerbated. As a result, models may converge to suboptimal solutions that rely on shortcuts \cite{geirhos2020shortcut}—non-causal features that achieve high ERM performance but fail to generalize. In high-stakes scientific settings, such failures undermine trustworthiness and replicability \cite{rudin2019stop,laine2021avoiding,kapoor2023leakage}.

Another challenge arises when IBs are misaligned with the scientific signal prior, thereby misdirecting the ERM optimization trajectory. A prominent example occurs in transfer learning, where models pre-trained on natural image datasets (e.g., ImageNet \cite{russakovsky2015imagenet}) are fine-tuned to mitigate overfitting in small-data regimes. While reuse of low-level features such as edges can sometimes accelerate convergence and improve generalization \cite{xie2018pre}, the initialization bias is tuned to natural image statistics rather than physicochemical signal structures, which can lead to degraded performance \cite{molnar2025unintended,raghu2019transfusion}. A related manifestation is the modality gap observed in multimodal fusion (e.g., RGB–thermal imaging); when using ImageNet-trained encoders, models systematically favor RGB channels due to their closer proximity to the model's initialization IB \cite{huang2021alleviating,zhou2025m}. To address such issues, scientific DL applications frequently introduce domain-aligned IBs, including physics-constrained losses \cite{cuomo2022scientific} or architectural modifications \cite{manifold2021versatile}. These approaches often outperform generic DL vision methods. Nevertheless, generic methods remain functional, albeit suboptimal, leading to a prevailing consensus that they constitute reasonable baselines, with domain alignment providing incremental gains. Consequently, many studies continue to adopt generic DL practices.

Here, we argue that the uncritical transposition of generic DL practices to scientific domains can lead not merely to suboptimality but to outright failure, even in the absence of an explicit initialization-related IB confound. We hypothesize that ostensibly ``generic'' practices are not truly modality-agnostic but instead implicitly human-centric, arising from a form of survival bias: research incentives systematically reward methods that perform well on natural image benchmarks, creating a feedback loop in which dominant IBs increasingly drift toward human RGB perception. This is further reinforced by biologically inspired components, such as CNNs, attention mechanisms \cite{vaswani2017attention}, and sparse coding \cite{papyan2017convolutional} that draw conceptual motivation from the human visual cortex. Consequently, the naive adoption of such methods in scientific imaging may inadvertently re-bias pattern recognition toward human optics, potentially suppressing the non-canonical information that the sensing modality was engineered to capture. Although shared priors and feature reuse can in some cases enable successful transfer, the converse is equally plausible: misalignment between IBs and data characteristics can produce severe performance degradation. We further contend that such failures are more pervasive than commonly acknowledged, often obscured by nominal gains over random or weak baselines, as well as by confirmation and publication biases that create a misleading appearance of success \cite{mcgreivy2024weak,saidi2025unraveling}.

Here, we illustrate these ideas by examining the case of infrared (IR) histopathological imaging, which seeks to be more accurate and informative than conventional pathology based on RGB digitization of stained tissues (e.g., hematoxylin and eosin, H\&E). It serves as an effective test bed for four reasons: (1) Signal complexity: Each pixel in IR imaging is a high-dimensional spectral feature that quantitatively measures absorbance from distinct vibrational modes of biomolecules—including proteins, nucleic acids, and lipids—enabling detailed characterization of tumors and their microenvironment \cite{fernandez2005infrared,bhargava2023digital}. This yields pathological information beyond standard H\&E staining, exemplifying a modality with rich information content and non-canonical contrast. (2) Experimental control: Because IR imaging is non-destructive, paired IR and H\&E images can be acquired from the same tissue section. This enables controlled, objective comparison of DL performance between human-centric (H\&E) and scientific (IR) modalities while minimizing dataset bias. (3) Foundational relevance: IR spectroscopy is a cornerstone of analytical chemistry \cite{haas2016advances}, so insights from this case study may generalize across a wide range of scientific applications. (4) High-stakes utility: Silent DL failures in pathology can lead to misdiagnosis or delayed treatment, making investigation of potential failure modes particularly urgent.

\begin{table}[t]
\centering
\fontsize{7.5pt}{8.5pt}\selectfont 
\renewcommand{\arraystretch}{1} 
\setlength{\tabcolsep}{0pt}      
\begin{tabularx}{\columnwidth}{l @{\extracolsep{\fill}} rrr rrr rrr}
\toprule
& \multicolumn{3}{c}{Accuracy (\%, $\uparrow$)} & \multicolumn{3}{c}{AUROC (\%, $\uparrow$)} & \multicolumn{3}{c}{AUPRC (\%, $\uparrow$)} \\
\cmidrule(lr){2-4} \cmidrule(lr){5-7} \cmidrule(lr){8-10}
Domain & \multicolumn{1}{c}{Train} & \multicolumn{1}{c}{Test} & \multicolumn{1}{c}{Gap} & \multicolumn{1}{c}{Train} & \multicolumn{1}{c}{Test} & \multicolumn{1}{c}{Gap} & \multicolumn{1}{c}{Train} & \multicolumn{1}{c}{Test} & \multicolumn{1}{c}{Gap} \\
\midrule
H\&E         & 90.1{\tiny$\pm$1.2} & 83.9{\tiny$\pm$3.9} & $-$6.2{\tiny$\pm$4.4}  & 95.8{\tiny$\pm$0.3} & 90.8{\tiny$\pm$1.4} & $-$5.0{\tiny$\pm$1.4} & 61.5{\tiny$\pm$2.0} & 52.2{\tiny$\pm$5.9} & $-$9.3{\tiny$\pm$7.3} \\
IR           & 87.6{\tiny$\pm$2.1} & 76.4{\tiny$\pm$5.0} & $-$11.2{\tiny$\pm$5.6} & 94.4{\tiny$\pm$0.9} & 83.0{\tiny$\pm$2.9} & $-$11.5{\tiny$\pm$3.2} & 59.4{\tiny$\pm$2.4} & 42.7{\tiny$\pm$6.5} & $-$16.8{\tiny$\pm$8.3} \\
Virtual H\&E & 89.5{\tiny$\pm$1.3} & 83.2{\tiny$\pm$4.0} & $-$6.4{\tiny$\pm$4.4}  & 95.5{\tiny$\pm$0.4} & 89.4{\tiny$\pm$1.9} & $-$6.1{\tiny$\pm$2.0} & 61.1{\tiny$\pm$1.9} & 48.5{\tiny$\pm$6.3} & $-$12.6{\tiny$\pm$7.3} \\
\bottomrule
\end{tabularx}
\caption{\textbf{Classification performance.} Mean $\pm$ 95\% CI. Accuracy, AUROC, and AUPRC are reported; ``gap'' indicates the difference between train and test sets.}
\label{tab1}
\end{table}

\begin{figure}[t!]
\centering
\includegraphics[width=\textwidth]{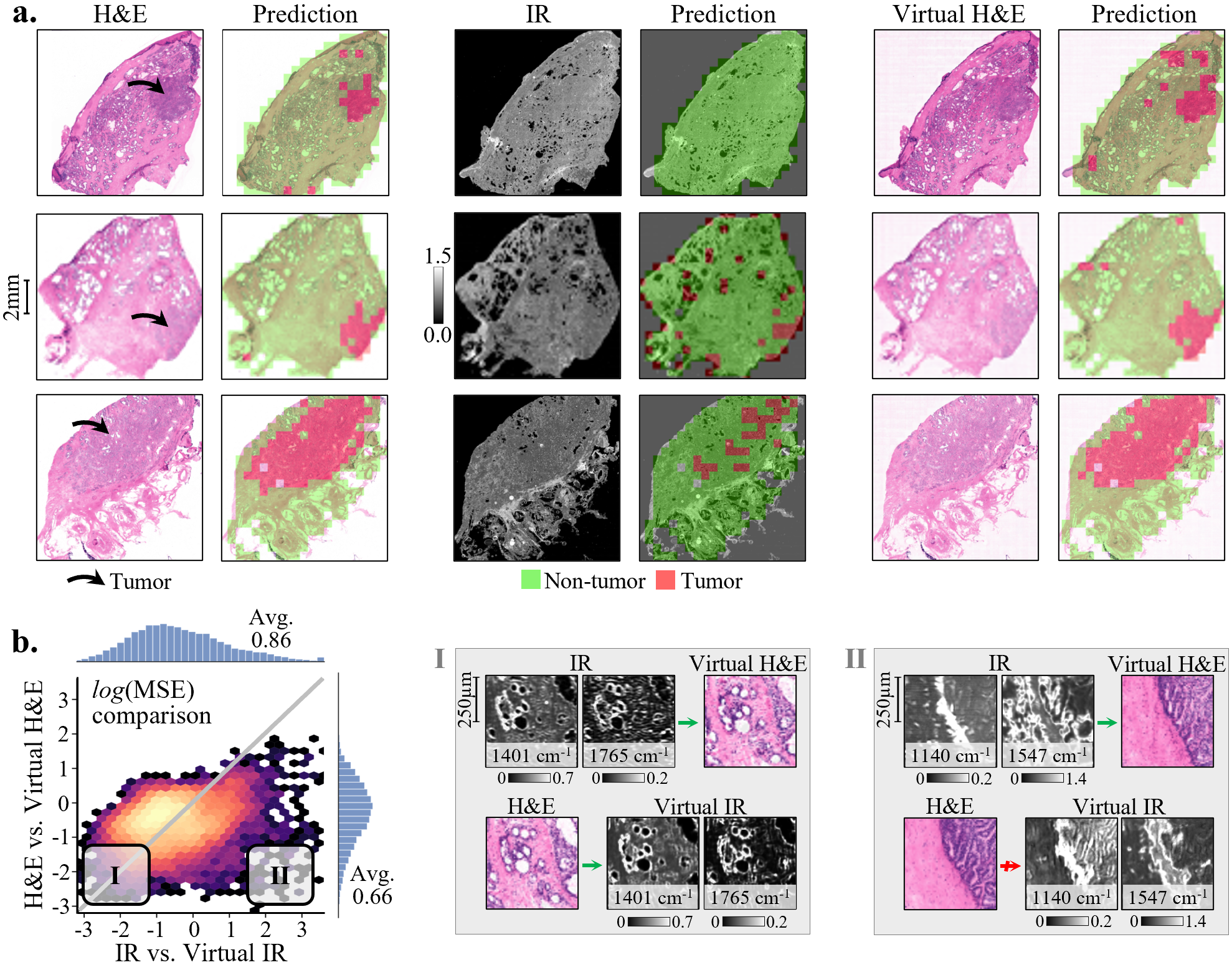}
\caption{\textbf{High-level comparison}. \textbf{a.} Example tile-level classification across input domains. \textbf{b.} Histogram comparing $\log(\text{MSE})$ between virtual and real test images in IR and H\&E domains. Note, data are standardized before MSE computation; lower MSE indicates higher translation accuracy. The right panel shows cases where: (I) IR and H\&E are mutually translatable, and (II) IR-to-H\&E translation is feasible, but the reverse is not.}
\label{fig1}
\end{figure} 

We begin by comparing DL vision models for tumor detection trained on matched IR and H\&E images under standard ERM. We observe that IR paradoxically underperforms H\&E, despite containing sufficient contextual information and irrespective of input dimensionality, thereby establishing a clear failure case. Importantly, this issue would remain hidden if IR performance were assessed in isolation, as the model still achieves acceptably high accuracy. We conduct a series of behavioral and mechanistic analyses to determine why and how this failure occurs, finding evidence that it stems from a mismatch between DL's IB and IR signal priors. We further examine the broader ramifications of this failure, particularly in edge clinical cases, trustworthy AI, and the role of advanced scientific imaging modalities such as IR. Finally, we benchmark a wide range of generic DL robustification methods to assess how well current approaches address this failure, extending our study from standard baselines to the state-of-the-art. Taken together, our work serves as a concrete counterexample to the assumption that generic DL practices transfer to scientific domains, and a call for more deliberate, modality-aware adaptation. 

\section{Results}
\subsection{IR-trained models underperform H\&E-trained models despite richer information} \label{s21}

\textit{Experimental setup.} To facilitate our study, we imaged prostate tissue sections using a laser scanning confocal IR microscope \cite{yeh2023infrared}, utilizing 10 mid-IR bands (1000--1800~cm$^{-1}$) as discrete image channels. To establish a gold-standard clinical reference aligned with human perception, the same sections were stained with the most commonly used pathology dye (H\&E), digitized in RGB, and registered to the IR slides. Both IR and H\&E slides were downsampled, tiled, and Gaussian-standardized to ensure distributional parity; this decouples low-level intensity statistics. Tiles were labeled ``tumorous'' or ``non-tumorous,'' representing a standard digital pathology task. We employed a patient-wise train/test split across 15 random folds and report average results. 

We conduct our primary analysis using a baseline DL setting: a ResNet50 \cite{he2016deep} CNN classifier trained from scratch via ERM. ResNet50 was selected as it remains a robust, routine choice among practitioners and frequently serves as the architectural backbone for exploratory studies. Unless otherwise noted, reported results correspond to this baseline; we explore other DL practices and alternative architectures in \S\ref{s26}.

\textit{Performance comparison.} We first evaluate models trained directly on H\&E and IR domains using three metrics: accuracy, area under the receiver operating characteristic curve (AUROC), and area under the precision-recall curve (AUPRC). As shown in Tab. \ref{tab1}, IR models underperform their H\&E counterparts on the test set, exhibiting a wider generalization gap. Specifically, the performance drop from train to test sets is more pronounced in the IR domain (Accuracy: $-5.0\%$, AUROC: $-6.4\%$, AUPRC: $-7.5\%$ relative to H\&E), suggesting greater overfitting to non-causal, spurious features. Qualitative assessment of whole-slide images in Fig. \ref{fig1}-a confirms these failures, manifesting as both complete false negatives in tumorous regions (rows 1 and 3) and false positives in non-tumorous regions (row 2). In diagnostic screening pipelines, such errors could lead to misdiagnosis or an inaccurate assessment of cancer progression, ultimately delaying or compromising treatment plans. 

\textit{Sanity check through virtual staining.} Three potential factors could trigger this failure: (1) insufficient informational content due to suboptimal IR system design (e.g., resolution limits, artifacts, or poor band selection), (2) limited model capacity, or (3) high-dimensionality issues, given that IR utilizes ten channels compared to H\&E’s three. We address the latter two points later. To address the first, we perform image translation between the H\&E and IR domains to assess mutual information. Per the Data Processing Inequality (DPI) \cite{yu2020understanding,chang2022explaining,liu2025information}, a neural network conditioned to input cannot generate new information; thus, successful translation serves as a proxy for information availability.

As shown in Fig. \ref{fig1}-b, while some images were mutually translatable, a distinct asymmetry emerged: IR images were reliably converted to H\&E, but the reverse was less successful. H\&E-to-IR translation was highly variable, succeeding in distinct morphologies but failing in ambiguous regions where IR’s spectral signatures are required for disambiguation. Conversely, IR-to-H\&E translation was consistently accurate, with the mean squared error (MSE) approximately 0.20 lower than the reverse. This implies IR encompasses more of H\&E’s content than vice versa. While some nuances may be lost, classification performance when trained on the virtual H\&E domain nearly matched that of real H\&E (Tab. \ref{tab1} and Fig. \ref{fig1}-a), indicating that these virtual images are diagnostically effective. Thus, an IR model should, in principle, be able to extract at least virtual H\&E-equivalent features to match this performance baseline---if not surpass it by leveraging IR-unique cues. The failure to do so points to a fundamental learning bottleneck rather than a lack of signal.

\subsection{IR-trained CNNs exhibit poor spectral-spatial learning, regressing toward 1D spectral analysis}\label{s22}

To characterize this failure, we analyze it from three angles.

\begin{figure}[t!]
\centering
\includegraphics[width=\textwidth]{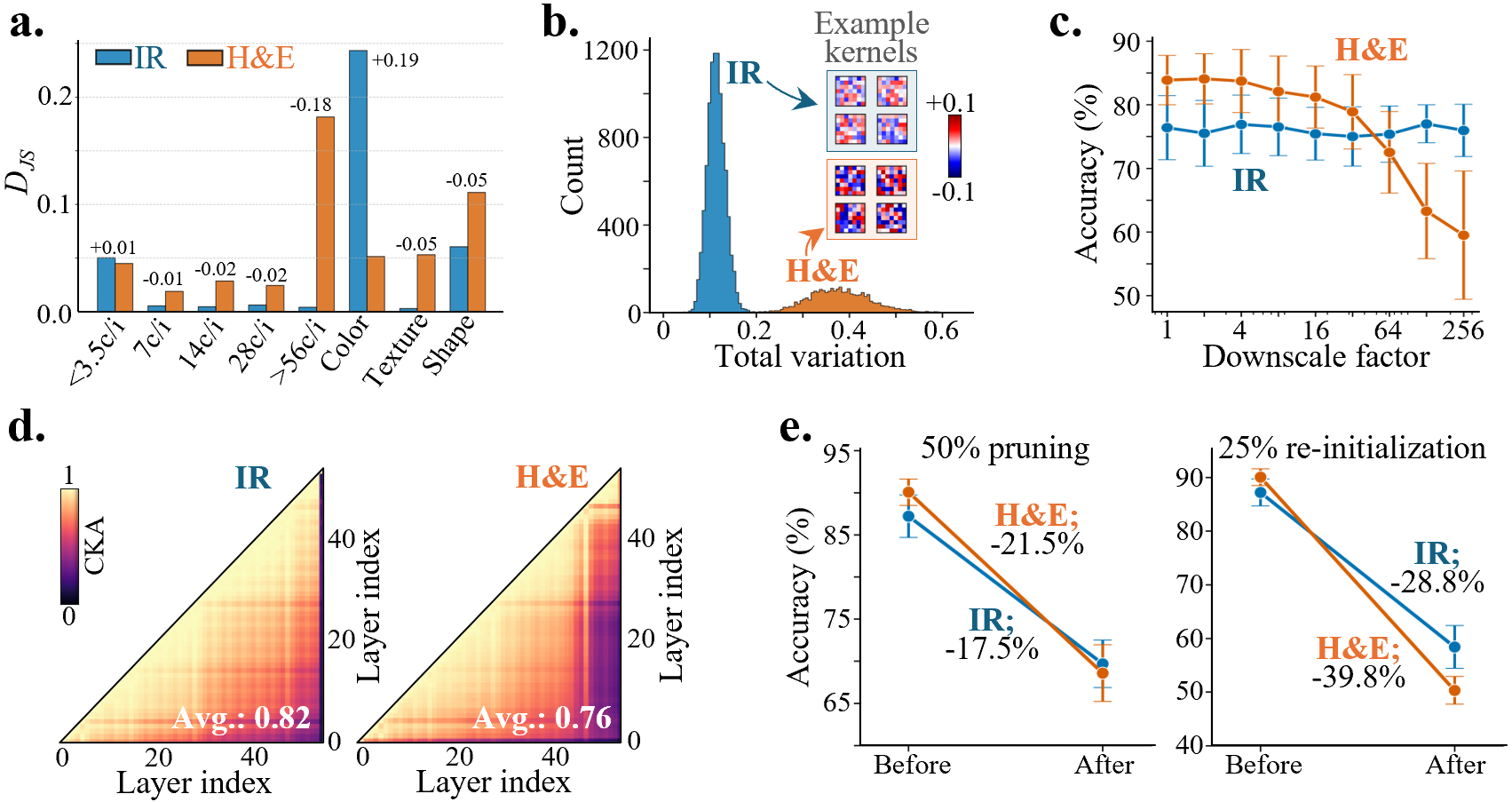}
\caption{\textbf{Cue analysis and network dissection.}
\textbf{a.} Sensitivity ($D_{JS}$) to spatial frequency and HVS cues, with excess sensitivity (IR – H\&E) shown above bars. 
\textbf{b.} Histogram of first-layer kernel total variation. 
\textbf{c.} Test accuracy across spatial downscaling factors; 1 denotes original resolution, 256 denotes full collapse. 
\textbf{d.} Intra-CKA computed on a test subset across all $\binom{54}{2}$ ResNet50 layer pairs. 
\textbf{e.} Test accuracy drop following layer pruning and reinitialization.
}
\label{fig2}
\end{figure} 

\textit{1. Cue manipulation:} To probe the model’s decision process, we perturb or mask visual cues in test images and quantify the resulting prediction changes using the Jensen–Shannon divergence ($D_{JS}$). A higher $D_{JS}$ indicates greater sensitivity and thus stronger reliance on the perturbed cue. Interventions are applied along two axes: {spatial frequencies}, where, following prior work \cite{solomon1994visual,yin2019fourier,li2023robust,subramanian2024spatial}, we selectively suppress frequencies across seven octave ranges (1.75–112 cycles/image, c/i); and {human visual system (HVS) cues}, where, following \cite{geirhos2018generalisation,ge2022contributions,hermann2020origins,mummadidoes}, we apply distortions targeting color (spectral), texture, and shape via channel jitter, blur, and grid shuffling operations, respectively, perturbing one cue at a time while keeping the others relatively intact.

As shown in Fig. \ref{fig2}-a, in the spatial-frequency domain, H\&E models peak at $>$56 c/i, accounting for 51\% of the total response ($D_{JS}/\Sigma D_{JS}$). This window captures high spatial frequency (HSF) patterns such as textures and edges, which support shape learning and higher-level abstraction. Other frequency ranges also elicit moderate responses, resulting in a relatively balanced sensitivity profile. In contrast, IR models rely predominantly on low spatial frequencies (LSF), with over 70\% of their response below 3.5 c/i, corresponding visually to smooth spectral gradients and broad structural variations. Their responses in other ranges, including HSF, are minimal. In the HVS-cue domain, H\&E models likewise exhibit a balanced profile, with each cue contributing $>$20\% of total response. IR models, however, are dominated by color (79\%), followed by shape (20\%) and texture (1\%). Given their limited HSF sensitivity, this shape processing likely occurs at a coarse level, with no edge awareness. These trends are further corroborated by early-layer kernel statistics: IR kernels are smoother and exhibit lower total variation (Fig. \ref{fig2}-b), indicating reduced encoding of HSF and texture. Overall, H\&E models exhibit heterogeneous, mixed cue utilization, whereas IR models are strongly concentrated on shallow spectral signatures such as color and LSF, despite the intrinsically richer signal content of IR images.

\textit{2. Spatial ablation:} We extend the above analysis by explicitly constraining image information, retraining models, and measuring performance changes. Because earlier results suggest IR models underutilize spatial information, we ablate spatial structure by downscaling (binning) images and then upsampling back to the original resolution, keeping input size constant while progressively removing spatial detail. We vary the downscaling factor up to the extreme case of full collapse into a 1D bulk spectra with no spatial context, where CNN training becomes effectively equivalent to an Multi-Layer Perceptron (MLP) on bulk spectra. The same procedure is applied to H\&E images for comparison. 

As shown in Fig. \ref{fig2}-c, H\&E models follow the expected trend: test accuracy declines monotonically with decreasing spatial resolution, and when reduced to a 1D RGB vector, performance drops substantially ($>$20\%). In contrast, IR models exhibit virtually no degradation ($<$0.5\%) even in the fully collapsed setting. To quantify model similarity, we compute Cohen’s $\kappa$ between predictions from the original-resolution and collapsed models. IR models show 84\% prediction overlap (including shared errors), yielding a moderate-to-strong $\kappa$ of 0.62, substantially higher than H\&E’s 0.26 and exceeding typical values reported in DL studies \cite{tuli2021convolutional,geirhos2020beyond}; this indicates strong behavioral equivalence from a test-metric standpoint. Although no test performance difference is observed, original-resolution IR models outperform collapsed counterparts on the training set by $\sim$5\% (not shown). This reflects improved ERM when spatial context is available and suggests that some spatial context is ``learned.'' However, because these gains do not transfer to test performance, their learned features are spurious, with limited generalizability.

\textit{3. Network dissection:} We further examine the hierarchical degree of spectral-spatial abstraction in original-resolution models. We use the Centered Kernel Alignment (CKA) metric \cite{kornblith2019similarity} to quantify similarity among hidden layer activations across depths within each model (intra-CKA). As shown in Fig. \ref{fig2}-d, layers in IR models remain highly similar throughout, indicating co-adaptation, redundancy, or both—hallmarks of poor hierarchical representation. Consistently, performance in IR models is less affected by unstructured $\ell_1$ pruning or random reinitialization of entire layers (Fig. \ref{fig2}-e), indicating that feature extraction across layers is degenerate or repeated. This trend may also reflect spurious localization, whereby causal and spurious features occupy separate subnetworks \cite{frankle2018lottery,qiu2024complexity}; because IR models exhibit stronger spurious tendencies (\S\ref{s21}), pruning or resetting may disproportionately disrupt spurious pathways, producing a partial robustification effect and smaller performance drops. In all cases, the intrinsic hierarchical structure of CNNs appears less effectively utilized in IR than in H\&E. This rules out limited model capacity as the source of failure, since IR models are, in fact, under-utilizing their available capacity.

Taken together, findings (1–3) converge on a consistent deficiency regarding IR models: they do not learn robust spatial—and, by extension, joint spectral-spatial—representations. Instead, their behavior approximates that of a 1D spectral analysis model, attaining reasonable accuracy largely due to the intrinsic discriminative power of IR spectra. This, however, is suboptimal, as it precludes integrated learning of the diverse biochemical signals and morphological cues present in IR images, which are essential for comprehensive pathological interpretation.\footnote{A common assumption is that biases toward shape and LSF promote generalization, whereas reliance on texture and HSF does not \cite{geirhosimagenet,wang2020high,li2023robust}. However, cue generalizability is task-dependent \cite{dai2022rethinking,ge2022contributions}. For instance, Qiu et al. \cite{qiu2024shape} showed that texture aids animal classification, whereas shape is more robust for artificial objects, consistent with cognitive science linking artificial object identity to function and thus shape. In histopathology, both LSF and HSF, along with all HVS cues, are diagnostically relevant, as reflected in classical computer-aided diagnostic systems that integrate diverse feature types \cite{mosquera2014computer,veta2014breast}.} Consequently, IR models generalize poorly and underperform H\&E model counterparts. Moreover, this regressive behavior represents a fundamental failure of vision-oriented DL methods, as the spatial IB motivating their use do not translate into enhanced performance.

\begin{figure}[t!]
\centering
\includegraphics[width=\textwidth]{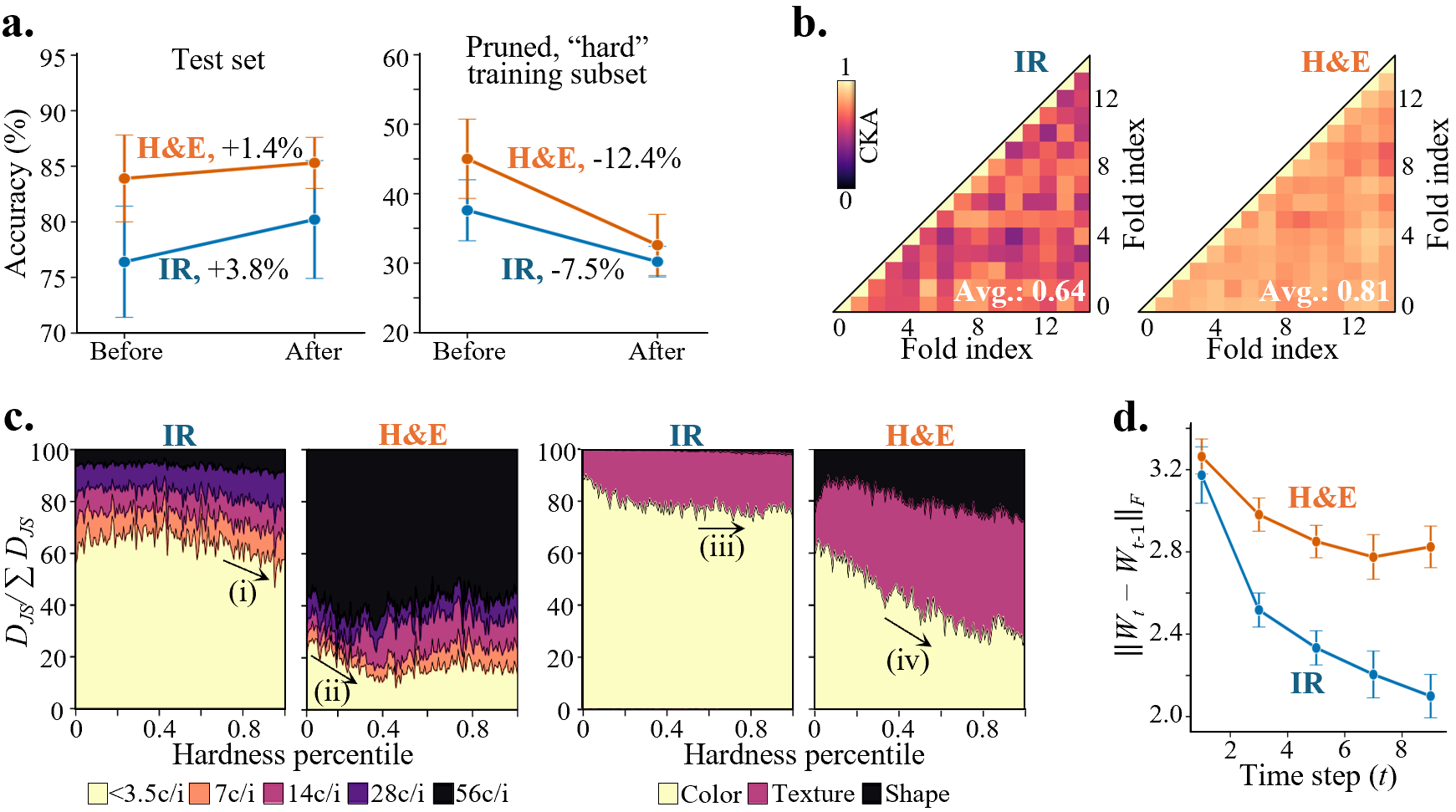}
\caption{\textbf{Overfitting modes and SB dynamics}. \textbf{a.} Accuracy measured before vs. after pruning the hard training subset, evaluated on both the test set and pruned subset. \textbf{b.} Inter-CKA computed on a subset of the combined train-test sets, using the last-layer activations across all $\binom{15}{2}$ model pairs trained on different train/test splits. \textbf{c.} Sensitivity ($D_{JS}$) responses to spatial frequency (left) and HVS cues (right) across sample hardness percentiles. \textbf{d.} Frobenius norm of the difference between CNN weights at consecutive time steps, evaluated every 100 iterations.}
\label{fig3}
\end{figure}

\subsection{IR models are more critically overfit}\label{s23}

\textit{Overfitting modes: distributed spuriousness vs. memorization.} Both H\&E and IR models overfit, with train–test accuracy gaps of $-6.2\%$ and $-11.2\%$, respectively (Tab. \ref{tab1}). Although IR’s poorer generalization is due to limited cue learning and thus poor spectral-spatial representations, their distinct cue sensitivity profiles additionally suggest subtle differences in overfitting mode. Overfitting may stem from {distributed spurious correlations}, where features persist across groups, or {sample-specific memorization}, where individual instances are interpolated. The former is “malignant,” as it interferes with core feature learning \cite{qiu2024complexity,zhang2024feature}, whereas memorization is comparatively “benign,” often coexisting with generalization \cite{belkin2019reconciling,d2020triple,mallinar2022benign,bayatpitfalls} and even viewed as necessary for capturing rare cases in long-tailed datasets \cite{feldman2020neural}. Because memorization is linked with HSF noise \cite{wang2020high}, we postulate that H\&E models—biased toward HSF—exhibit this milder form more, while LSF-dominated IR models rely more on harmful distributed spurious patterns that exacerbates test error.

To evaluate this, we estimate memorization extent using Feldman’s criterion \cite{feldman2020neural}, defining a sample as memorized if it is correctly classified during training but misclassified when withheld. As full leave-one-out retraining is infeasible, we adopt subset-based approximations \cite{jiang2021characterizing,lukasiklarger}, pruning the top 5\% ``hardest'' training samples, which are most likely to be memorized. We quantify hardness via a composite ranking combining convergence timing \cite{jiang2021characterizing,pleiss2020identifying,paul2021deep}, forgetting frequency \cite{tonevaempirical,maini2022characterizing}, gradient noise \cite{agarwal2022estimating}, minority group membership \cite{sorscher2022beyond,lee2018simple}, and computational complexity \cite{baldock2021deep}. As shown in Fig. \ref{fig3}-a, both IR and H\&E models exhibit memorization among difficult training examples. Note, models are not over-trained to zero loss, so hard sample accuracy remains low even when included in training. H\&E models show higher memorization (+4.9\%), consistent with their relatively stable test performance before and after pruning. In contrast, IR models display lower memorization, suggesting comparatively greater reliance on distributed spurious patterns; correspondingly, pruning yields a +2.36\% test accuracy gain from disrupting these patterns, supporting the view that IR models are more adversely overfit.

\textit{Algorithmic stability.} From the perspective of algorithmic stability \cite{bousquet2002stability}, these findings further suggest that learning in IR domains is comparatively less stable. The reliance on distributed spurious patterns leads to increased sensitivity to the empirical training distribution. This renders the effective solution space in IR modalities to be much broader and volatile, susceptible to dataset biases and complicating generalization. In contrast, H\&E models exhibit relatively higher stability: they tend to converge toward more consistent decision boundaries, with memorization acting as a passive secondary mechanism for handling dataset-specific outliers. To show this, we compute CKA between models trained on different train/test folds (inter-CKA). As shown in Fig. \ref{fig3}-b, IR models display lower mean inter-CKA scores, supporting the interpretation of reduced stability.

\subsection{Simplicity bias limits generalization in IR models}\label{s24}

\textit{Simplicity bias (SB).} DL models are well documented to exhibit a SB \cite{arpit2017closer,de2019random,valledeep,rahaman2019spectral,hu2020surprising,geirhos2020shortcut,shah2020pitfalls}—the tendency to favor salient low-complexity patterns over more complex ones. This hallmark property is a double-edged sword that both enables and constrains DL: it helps explain why overparameterized neural networks generalize on high-dimensional image data rather than overfitting to noise, defying classical bias–variance trade-off theory \cite{de2019random,valledeep}, yet real-world decision boundaries are often highly complex, and SB can hinder their learning.

\textit{IR signal properties.} IR's spectral dimension exhibit three key properties: \textit{(P1) Easy-to-learn}, as features are directly accessible from raw inputs—a canonical property irrespective of modality, e.g., H\&E color; \textit{(P2) Class discriminativeness}, where spectral features alone partially satisfy ERM (mean training accuracy 82.6\%); and \textit{(P3) Causality}, where these features generalize reasonably to test data (mean accuracy 76.0\%). P2 and P3 reflect the corollaries of the biochemical contrast of IR and its direct relevance to histopathology. By comparison, H\&E color signals, while encoding chemical affinity (e.g., nuclei stain blue/purple, cytoplasm/stroma pink/red), are largely qualitative and lack these characteristics. 

\textit{IR prior-SB interactions.} We hypothesize that the observed failure arises from a detrimental interaction between SB and IR’s properties P1-3. The postulated mechanism is as follows: the interaction between P1 and SB leads models in both IR and H\&E to initially rely on spectral/color and other simple cues. In H\&E, these cues poorly satisfy ERM, incentivizing the development of spatial abstractions and progressively richer, generalizable features. In IR, however, P2 ensures most samples are already well fit, reducing pressure to escape SB and learn true spectral-spatial structure. To fit the remaining minority cases, models instead resort to lazy extensions of existing spectral representations (e.g., slight gradient or layout variations). These features may be spectral-spatial and even causal, but are minimally complex and tailored to few samples; however, their lazy nature makes them more likely to be spurious shortcuts. This outcome is SB-consistent, as learning new spectral-spatial abstractions is costlier while benefiting only a small data fraction. Prior works support this, where early-phase biases are difficult to reverse, often slowing or suppressing acquisition of more causal features \cite{achille2018critical,sagawa2020investigation,qiu2024complexity,yang2024identifying}. Consequently, training accuracy rises, but the weak robustness of the lazy features causes test performance to plateau, limiting generalization to what P3 supports.

\textit{Validation through sample hardness.} We test whether this narrative holds by examining whether easy training samples exhibit stronger reliance on SB-aligned attributes, such as color and LSF, reflecting early SB-dominated learning. We then assess whether cue-reliance trends diverge for harder samples, with H\&E models shifting toward additional features while IR models fail to do so. We adopt the same hardness metrics as before. As shown in Fig. \ref{fig3}-c, easy samples in both domains exhibit high reliance on color and LSF, though H\&E values are weaker, likely because the absence of IR-specific P2 properties incentivizes earlier exploration of alternative cues. For harder samples, a clear divergence emerges: H\&E models shift sensitivity toward other cues such as texture and HSF, whereas IR models show weaker or no transition. Along the spatial-frequency axis, LSF influence begins to diminish only in the upper quantile of hard samples in IR (i), compared to an earlier reduction in H\&E (ii). Across HVS cues, this effect is more pronounced: color progressively diminishes in importance with increasing hardness in H\&E (iv), but remains dominant in IR throughout (iii). To further support the presence of lazy learning in this regime, we measure CNN weight displacement over training (Fig. \ref{fig3}-d), finding lower changes in IR models than in H\&E.

\subsection{Failure is not caused by high dimensionality}\label{s25}
We test whether IR underperformance stems from its high Ambient Dimensionality (AD), which is $\sim$3.3× that of H\&E. High AD is linked to sparsity, redundancy, noise, and distance collapse issues, all of which can hinder generalization. The central question is whether the failure is due to dimensionality itself or to modality-specific signal priors (contrast mechanisms, feature distributions, and cue structure). To isolate AD, we construct three dimension-reduced IR variants using Principal Component Analysis (PCA), an Autoencoder (AE), and a supervised MLP feature extractor, compressing the 10 IR bands to 3 components to match H\&E AD. If high AD were the primary cause, performance should improve. Instead, Tab. \ref{tab2} shows that accuracy does not recover and often degrades. Cue manipulation analysis further shows that all variants remain weakly sensitive to HSF and texture—the core deficiency of the original IR model—still far below H\&E. Although HSF sensitivity increases slightly, likely because loss of task-relevant spectral information incentivizes spatial learning, the shift is insufficient to improve performance. Thus, these variants remain susceptible to the same deficient learning mechanism, resulting in similar underperformance.

\begin{table}[t]
\centering
\fontsize{7.5pt}{8.5pt}\selectfont 
\renewcommand{\arraystretch}{1} 
\setlength{\tabcolsep}{0pt}      
\begin{tabularx}{\columnwidth}{l @{\extracolsep{\fill}} rrr rrr rrr}
\toprule
& \multicolumn{4}{c}{Dimensionality} & \multicolumn{2}{c}{$D_{JS} / \Sigma D_{JS}$ (\%)} & \multicolumn{3}{c}{Accuracy (\%)} \\
\cmidrule(lr){2-5} \cmidrule(lr){6-7} \cmidrule(lr){8-10}
Domain & \multicolumn{1}{c}{AD} & \multicolumn{1}{c}{ED (LPCA)} & \multicolumn{1}{c}{ED (MLE)} & \multicolumn{1}{c}{ED (2NN)} & \multicolumn{1}{c}{HSF} & \multicolumn{1}{c}{Texture} & \multicolumn{1}{c}{Train} & \multicolumn{1}{c}{Test} & \multicolumn{1}{c}{Gap} \\
\midrule
H\&E         & 151 K & 17.7{\tiny$\pm$0.1} & 31.4{\tiny$\pm$1.8}  & 63.3{\tiny$\pm$4.2} & 57.5{\tiny$\pm$4.3} & 24.8{\tiny$\pm$1.5} & 90.1{\tiny$\pm$1.2} & 83.9{\tiny$\pm$3.9} & $-$6.2{\tiny$\pm$4.4} \\

IR           & 502 K & 17.1{\tiny$\pm$0.1} & 19.9{\tiny$\pm$1.1} & 43.8{\tiny$\pm$2.4} & 8.3{\tiny$\pm$1.6} & 1.7{\tiny$\pm$0.3} & 87.6{\tiny$\pm$2.1} & 76.4{\tiny$\pm$5.0} & $-$11.2{\tiny$\pm$5.6} \\

Virtual H\&E & 151 K & 17.6{\tiny$\pm$0.1} & 38.1{\tiny$\pm$2.9}  & 57.7{\tiny$\pm$3.9} & 56.2{\tiny$\pm$4.5} & 33.1{\tiny$\pm$6.7} & 89.5{\tiny$\pm$1.3} & 83.2{\tiny$\pm$4.0} & $-$6.4{\tiny$\pm$4.4} \\

IR (PCA) & 151 K & 17.2{\tiny$\pm$0.1} & 19.5{\tiny$\pm$1.5}  & 44.7{\tiny$\pm$2.5} & 11.9{\tiny$\pm$3.5} & 6.9{\tiny$\pm$1.1} & 83.9{\tiny$\pm$1.4} & 69.6{\tiny$\pm$6.3} & $-$14.3{\tiny$\pm$6.9} \\

IR (AE) & 151 K & 16.9{\tiny$\pm$0.2} & 19.8{\tiny$\pm$0.8}  & 37.3{\tiny$\pm$3.4} & 21.4{\tiny$\pm$3.2} & 9.3{\tiny$\pm$1.6} & 82.6{\tiny$\pm$1.0} & 68.5{\tiny$\pm$6.8} & $-$14.1{\tiny$\pm$7.5} \\

IR (MLP) & 151 K & 17.2{\tiny$\pm$0.1} & 24.8{\tiny$\pm$2.0}  & 48.4{\tiny$\pm$5.5} & 27.6{\tiny$\pm$5.0} & 7.9{\tiny$\pm$1.6} & 86.1{\tiny$\pm$1.3} & 74.3{\tiny$\pm$5.2} & $-$11.8{\tiny$\pm$5.7} \\
\bottomrule
\end{tabularx}
\caption{\textbf{Dimension-reduced experiments.} Mean $\pm$ 95\% CI of AD, ED (based on different ID estimates on the DL model's last representation space), select sensitivity responses, and accuracy are reported.}
\label{tab2}
\end{table}

To better understand this, we distinguish Intrinsic Dimensionality (ID), the minimal degrees of freedom needed to describe the data, from AD. Although images have high AD, CNNs generalize because data lie on low-ID manifolds that networks implicitly exploit; probing the learned representation's ID reveals the Effective Dimensionality (ED) of the task-relevant subspace. A common view is that high AD exacerbates underspecification and noise fitting, yielding high-ED, noisy manifolds \cite{ansuini2019intrinsic,papyan2020prevalence}. Since ED is upper-bounded by AD,\footnote{Dimensionality hierarchy: ED $\lesssim$ ID $\ll$ AD.} dimension reduction regularizes against such overfitting. However, ED estimates based on Local PCA \cite{fukunaga1971algorithm}, Maximum Likelihood Estimation (MLE) \cite{levina2004maximum}, Two Nearest Neighbor (2NN) \cite{facco2017estimating} show that IR-centric models actually have lower ED than H\&E models. The failure therefore lies in the opposite regime: they collapse to overly low ED, consistent with SB dominance. This issue arises independently of AD, as both high and reduced (but still high) AD permit ultra-low ED collapse, evidenced by the similarly low ED observed in both the original and dimension-reduced IR models. Dimension reduction offers little benefit if it retains the IR-specific properties (P1–P2) that drive SB and subsequent low-ED collapse. This explains why naive reductions (PCA, AE, MLP) fail: spectral variance (P2) is largely preserved, while compression simplifies the signal (P1), making it at least as easy for DL models. The problem is therefore not dimensional but rooted in signal priors and their DL interaction. This explains why certain reduced forms, such as IR-translated virtual H\&E, avoid this issue—not because of lower dimensionality per se, but because the transformation reshapes signal structure towards that of H\&E, disrupting these properties but retaining diagnostic content.

\begin{figure}[t!]
\centering
\includegraphics[width=\textwidth]{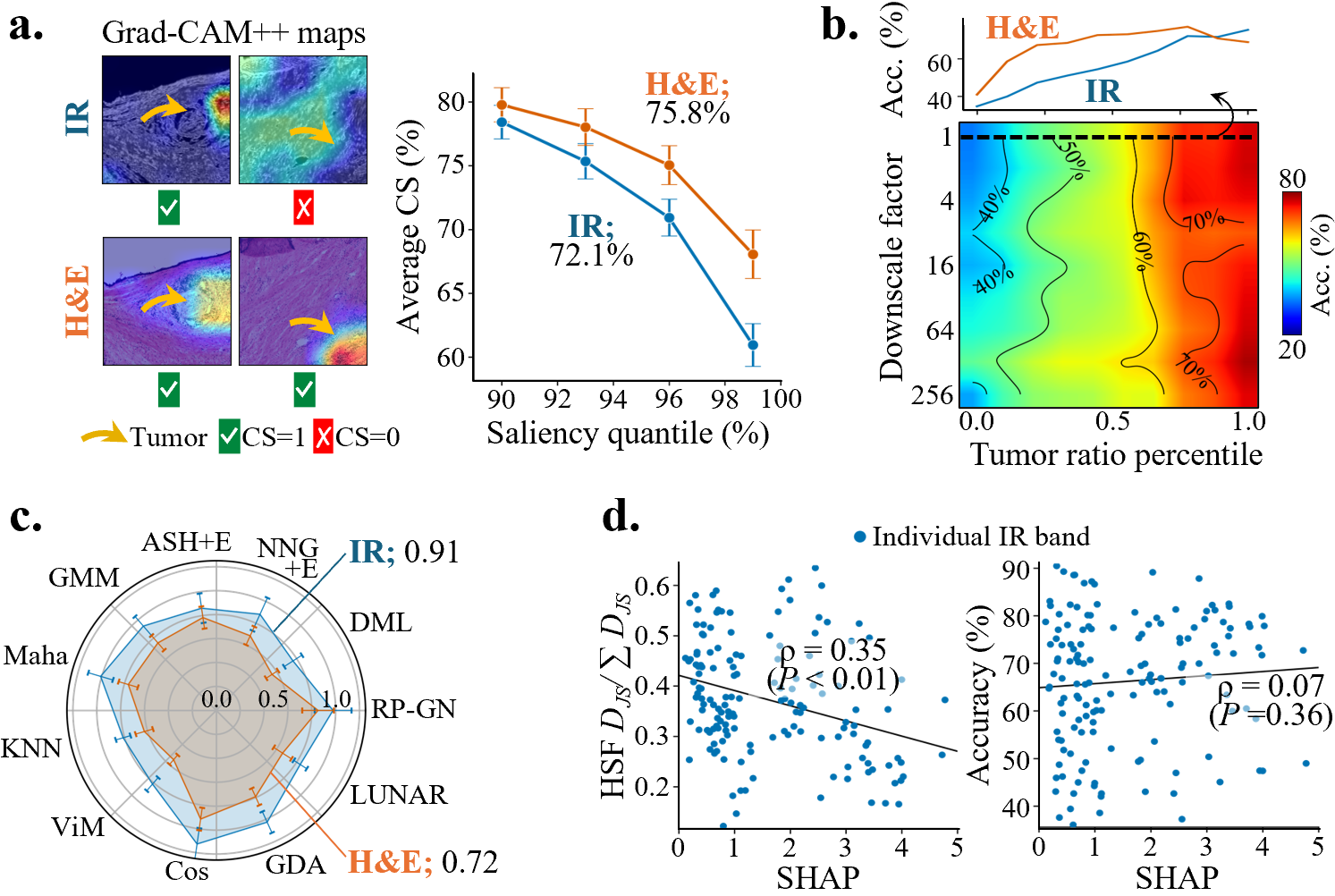}
\caption{\textbf{Failure repercussions.}
\textbf{a.} Grad-CAM++ saliency maps overlaid with ground-truth tumor regions (left), with CS curves (right) computed by thresholding the top 1–10\% of saliency as attended regions.
\textbf{b.} Test accuracy across tumor-ratio percentiles and spatial downscaling factors.
\textbf{c.} For 11 EU estimators, normalized ECE after progressively rejecting the top 1–90\% highest-EU test samples. Lower is better; 1 denotes random rejection.
\textbf{d.} Across IR bands, spectral usefulness (SHAP importance from fully collapsed IR models) vs. HSF response and test accuracy of the corresponding single-band CNN.}
\label{fig4}
\end{figure}

\subsection{Broader implications of the failure}\label{s26}
Earlier sections showed that this failure degrades generalization (e.g., tumor localization precision) and computational efficiency, as added model complexity provides little benefit for DL vision systems. If unaddressed, it also weakens the rationale for IR \textit{imaging}, as its spatial resolution—obtained at the expense of higher latency and memory demands, critical for whole-slide imaging—becomes unnecessary, since bulk spectral measurements at much lower resolution would suffice. These limitations impede clinical deployment, particularly in time- or resource-constrained settings such as intraoperative, on-device, or point-of-care use. Here, we extend the discussion.

\textit{Tumor localization.} HSF supports early edge extraction and downstream shape semantics, yet IR models underutilize it. We hypothesize this impairs robust tumor localization in cancerous tiles, which we test with two experiments.

1. Explainability validation: We apply Grad-CAM++ \cite{chattopadhay2018grad} to generate saliency maps on cancerous tiles and assess whether models attend to tumor regions. Using refined pixel-level tumor masks, we define a Concordance Score (CS) based on overlap between the mask and a top-percentile thresholded saliency map (CS=1 if any overlap, else 0). As shown in Fig. \ref{fig4}-a, IR models yield lower CS than H\&E, indicating predictions more often rely on regions containing no tumor pixels. Although microenvironmental context can be informative, direct tumor localization remains the most causal and clinically accepted diagnostic basis. Thus, such behavior, regardless of prediction correctness, undermines trustworthiness.

2. Small tumor blindness: The localization challenge is particularly acute for small objects \cite{achanta2009frequency,shi2025hs}; here, this includes tumors with small radii as well as larger tumors occupying only a small fraction of a tile’s field of view. Such tumors cannot be reliably detected by fully collapsed bulk-spectra models, as their signals are diluted through averaging. Because IR models exhibit functionally similar behavior, they are expected to inherit this same limitation. We investigate this in Fig. \ref{fig1}-b, which plots accuracy against tumor-to-tile area ratio across spatial downscaling factors from \S\ref{s22}. We confirm that IR models show markedly poorer detection than H\&E at low ratios, a trend consistent across all downscaling levels. In the broader histopathology context, this poses serious robustness concerns, as many diagnostic tasks involve sparse tumor signals and “needle-in-a-haystack” scenarios, e.g., micrometastasis detection \cite{bejnordi2017diagnostic} and microsatellite instability prediction \cite{kather2019deep}.

\textit{Uncertainty calibration.} Open-world deployment requires uncertainty quantification (UQ), particularly in high-stakes digital pathology, to flag errors preemptively and guide risk-aware intelligent decision-making \cite{lambert2024trustworthy,oh2025finer,nath2020diminishing}. Models must therefore produce well-calibrated, source-disentangled uncertainties. Using Deterministic UQ (DUQ) methods as a strong and efficient baseline \cite{mucsanyi2024benchmarking,oh2024we}, we extend the IR vs. H\&E comparison to evaluate which model offers more reliable uncertainty estimates.

1. {Aleatoric Uncertainty (AU):} AU reflects irreducible data uncertainty, such as class overlap, noise, or label disagreement, which sets the Bayes error floor. In DUQ, AU is assessed via softmax calibration—e.g., a prediction with 0.8 confidence should be correct 80\% of the time. Using Expected Calibration Error (ECE) \cite{guo2017calibration}, IR models show worse calibration (0.16) than H\&E models (0.11), indicating less reliable AU estimates.

2. {Epistemic Uncertainty (EU):} EU captures model uncertainty, i.e., the learner’s lack of knowledge, often arising under distribution shifts. In DUQ, EU is commonly estimated from feature distances (LUNAR \cite{goodge2022lunar}, GDA \cite{mukhoti2023deep}, GMM \cite{ahuja2019probabilistic}, Mahalanobis \cite{lee2018simple}, cosine \cite{techapanurak2020hyperparameter}, KNN \cite{sun2022out}), logits (DML \cite{zhang2023decoupling}), joint feature-logit signals (ASH+Energy \cite{liu2020energy,djurisicextremely}, NNGuide \cite{park2023nearest}, ViM \cite{wang2022vim}), or gradients (RP-GradNorm \cite{huang2021importance,jiang2023detecting}). High-EU samples often make both predictions and AU unreliable \cite{valdenegro2022deeper,wimmer2023quantifying}, motivating abstention. We evaluate EU calibration by progressively rejecting high-EU samples and computing normalized ECE on the retained set \cite{mucsanyi2024benchmarking}, with lower post-rejection ECE indicating better calibration. As shown in Fig. \ref{fig4}-c, IR models consistently underperform across all EU estimators. This likely arises because high-EU histopathology cases involve subtle structural shifts (e.g., gland morphology, stromal density) with similar low-level appearances, such as color. IR models’ limited cue use and spatial collapse render these shifts weakly expressed in feature and logit spaces, producing systematic overconfidence.

\textit{Spectral-spatial dilemmas.} A core advantage of IR imaging is the fusion of spectroscopy and optical microscopy, enabling spectral-spatial acquisition. In practice, we select the most interpretable (biologically grounded) and task-relevant spectral bands, expecting additional gains once spatial context is incorporated, ideally approaching an “oracle” regime. However, our narrative in \S\ref{s24} suggests a fundamental trade-off that obstructs this: the more “useful” a spectral band is—i.e., the more strongly it exhibits P1–P3—the less spatial learning occurs, preventing true joint spectral-spatial learning. We show this effect by training models on individual IR bands, each differing in spectral usefulness. Usefulness is quantified using Deep SHAP \cite{lundberg2017unified} on the supervised MLP in \S\ref{s25}, capturing only spectral learning without spatial context; note, these values vary across bands but also slightly across train/test splits. Fig. \ref{fig4}-d plots each band’s SHAP value against its standalone model’s HSF sensitivity and test accuracy. We observe that more useful bands show weaker HSF sensitivity, with a statistically significant negative correlation. This trend holds despite Deep SHAP’s limits in capturing higher-order, joint relations and possible variation in intrinsic spatial relevance across bands. Meanwhile, test accuracy shows no significant correlation with SHAP. Together, these results support the trade-off: highly useful spectral bands suppress spatial learning, whereas less useful bands induce it, yielding similar performance through divergent strategies. Thus, the theoretical advantages of IR imaging can be underutilized in DL vision, motivating mitigation strategies to help translate them into practical gains.

\subsection{Mainstream robustification strategies offer limited gain}\label{s27}
We move beyond the vanilla ERM ResNet50 baseline to evaluate more advanced, state-of-the-art DL practices, focusing on robustification methods drawn broadly from the literature. These approaches introduce diverse training-time interventions designed to steer the IB toward more generalizable solutions. While well validated on natural image benchmarks, they remain underexplored in scientific domains; here, we assess their effectiveness. We group methods by robustification principle, report results in Tab. \ref{tab3}, and interpret their performance in light of the mechanisms identified earlier.

\begin{table*}[t]
\centering
\fontsize{7.5pt}{8.5pt}\selectfont 
\renewcommand{\arraystretch}{1.05} 
\setlength{\tabcolsep}{2pt} 

\begin{minipage}[t]{0.47\linewidth}
\centering
\begin{tabularx}{\linewidth}{l @{\extracolsep{\fill}} r}
\toprule
Method & Accuracy \\
\midrule
H\&E & 83.9{\tiny$\pm$5.0} (+\color{blue}7.5)\\
Virtual H\&E & 83.2{\tiny$\pm$5.1} (+\color{violet}6.8)\\
Color jitter & 79.7{\tiny$\pm$5.2} (+\color{black}3.3)\\
Cutout & 78.4{\tiny$\pm$6.4} (+\color{black}2.0)\\
IRMv1 & 79.2{\tiny$\pm$6.6} (+\color{black}2.8)\\
V-REx & 79.1{\tiny$\pm$6.5} (+\color{black}2.7)\\
CVaR & 78.7{\tiny$\pm$6.2} (+\color{black}2.3)\\
JTT & 80.4{\tiny$\pm$5.1} (+\color{black}4.0)\\
LfF & 78.6{\tiny$\pm$6.2} (+\color{black}2.2)\\
SPARE & 78.8{\tiny$\pm$7.3} (+\color{black}2.4)\\
SiFeR & 80.0{\tiny$\pm$6.7} (+\color{black}3.6)\\
ReBias & 79.0{\tiny$\pm$6.3} (+\color{black}2.6)\\
ETD & 78.8{\tiny$\pm$5.9} (+\color{black}2.4)\\
Focal loss & 76.5{\tiny$\pm$6.5} (+\color{black}0.1)\\
\bottomrule
\end{tabularx}
\end{minipage}
\hfill
\begin{minipage}[t]{0.47\linewidth}
\centering
\begin{tabularx}{\linewidth}{l @{\extracolsep{\fill}} r}
\toprule
Method & Accuracy \\
\midrule
Mixup & 78.2{\tiny$\pm$7.0} (+\color{black}1.8)\\
MCD & 79.1{\tiny$\pm$7.3} (+\color{black}2.7)\\
HLC & 78.4{\tiny$\pm$6.5} (+\color{black}2.0)\\
M-heads + ESB & 79.6{\tiny$\pm$6.0} (+\color{black}3.2)\\
ConvNeXt & 78.9{\tiny$\pm$6.0} (+\color{black}2.5)\\
Swin v2 & 77.3{\tiny$\pm$8.5} (+\color{black}0.9)\\
MLP-mixer & 78.3{\tiny$\pm$6.8} (+\color{black}1.9)\\
Virtual H\&E finetune & 78.3{\tiny$\pm$6.9} (+\color{black}1.9)\\
Virtual H\&E $\ell_2$-SP & 79.1{\tiny$\pm$7.5} (+\color{black}2.7)\\
Virtual H\&E multitask & 83.2{\tiny$\pm$4.1} (+\color{violet}6.8)\\
Virtual H\&E early fusion & 83.2{\tiny$\pm$6.5} (+\color{violet}6.8)\\
Virtual H\&E late fusion & 84.7{\tiny$\pm$5.6} (+\color{blue}8.3)\\
Virtual H\&E late fusion + FiLM & 83.7{\tiny$\pm$6.6} (+\color{violet}7.3)\\
Virtual H\&E late fusion + CL & 83.6{\tiny$\pm$6.0} (+\color{violet}7.2)\\
\bottomrule
\end{tabularx}
\end{minipage}

\vspace{5pt}
\caption{\textbf{Mitigation benchmark.} Mean $\pm$ 95\% CI of test accuracy (excess relative to IR). H\&E (real and virtual) performance is shown for reference. Gains exceeding the virtual H\&E baseline are marked in violet, and those surpassing H\&E in blue.}
\label{tab3}
\end{table*}

\textit{1. Data augmentation and Invariant Risk Minimization (IRM):} Data augmentation promotes invariance and can disrupt spurious correlations. Since the core issue is over-reliance on spectral cues with weak spatial learning, we apply color jitter and Cutout \cite{devries2017improved}. Treating color-jittered samples as separate “environments,” we further evaluate IRM-based methods that penalize non-universal features across environments, using IRMv1 \cite{arjovsky2019invariant} and V-REx (Variance of training risk) \cite{krueger2021out}. These yield modest gains ($\sim$2.7\%) but remain well below H\&E (both real and IR-translated virtual) baselines. A likely reason is that spectral cues are partly causal (P3); indiscriminately penalizing them risks suppressing essential signal alongside spurious components.

\textit{2. Distributionally Robust Optimization (DRO) variants:} DRO focuses on worst-performing groups, typically by upweighting or upsampling training examples the model less focuses on, thereby counteracting bias. Here, these correspond to samples that deviate from the dominant spectral SB. We evaluate CVaR (Conditional Value-at-Risk) \cite{levy2020large}, JTT (Just Train Twice) \cite{liu2021just}, LfF (Learning-from-Failure) \cite{nam2020learning}, and SPARE (SePArate early and REsample) \cite{yang2024identifying}. Gains are again modest ($\sim$2.7\%) and remain below H\&E baselines. Although de-emphasizing SB-aligned samples is conceptually appropriate, its practical effectiveness is limited, likely because SB in IR is especially dominant, causing optimization to drift back toward them despite reweighting.

\textit{3. Feature debiasing:} SiFeR (SIeving Features for Robust learning) \cite{tiwari2023overcoming} and ReBias \cite{bahng2020learning} attempt feature-level debiasing by disentangling and suppressing early-layer spurious representations, but again yield limited gains ($\sim$3.1\%). Their suppression primarily targets color and other shallow cues, which, although often spurious in natural images, can be causal in our setting (P3), creating the same tension seen with augmentation. ETD (Example-tied Dropout) \cite{maini2023can} localizes spurious signals to specific neurons and deactivates them at test time, but achieves similarly small improvements (2.4\%). While ETD avoids the spurious–causal conflict, it is mainly validated for memorization-type spuriousness, whereas our failure mode is more distributed, limiting its effectiveness.

\textit{4. Robust losses and uncertainty quantification (UQ):} These general-purpose overfitting countermeasures aim to improve representation robustness and calibration, which could potentially alleviate the failure. We evaluate Focal Loss \cite{lin2017focal}, Mixup \cite{zhang2018mixup}, MCD (Monte Carlo Dropout) \cite{gal2016dropout}, Heteroscedastic Logit Classifier (HLC) \cite{kendall2017uncertainties}, and a Multi-head Ensemble \cite{lee2015m} with ESB (Evading Simplicity Bias) diversity regularization \cite{teney2022evading}. Gains are again limited ($\sim$2\%), slightly below prior categories. This is expected, as they do not explicitly target bias, instead promoting smoother decision boundaries, reduced overconfidence, or model averaging. These mechanisms are most effective against overfitting to sharp, high-frequency boundaries, whereas our issue is the opposite: IR models under-learn such boundaries.

\textit{5. Architectural upgrades:} Modern architectures increasingly hard-code IB that favor richer spatial reasoning, including ConvNeXt \cite{liu2022convnet}, Swin Transformer \cite{liu2022swin}, and MLP-Mixer \cite{tolstikhin2021mlp}. Prior studies suggest they capture global structure and long-range dependencies more effectively, even without large-scale pretraining, improving robustness and generalization \cite{bai2021transformers,paul2022vision,benz2021adversarial,bhojanapalli2021understanding,naseer2021intriguing,pinto2022impartial}. We replace ResNet50 with capacity-matched variants of these models, observing limited gains ($\sim$1.8\%). One reason is that low-level spectral signatures remain easily exploitable in early stages (e.g., patch embeddings in Swin and MLP-Mixer, and convolutions in ConvNeXt). Thus, although they encourage more expressive deep representations, they remain vulnerable to spectral SB.

Together, these methods yield limited improvements because they are misaligned with our specific failure mechanism, precipitated through the initial spectral (color) SB that confines learning. In contrast, most were designed for natural-image settings where SB manifests as reliance on color, texture, or background shortcuts (e.g., firetrucks with red paint, cows with grass) \cite{beery2018recognition,geirhosimagenet,baker2018deep}. Although both stem from SB, two distinctions matter: (1) causality and (2) severity. Here, spectral SB corresponds to partially causal learning (P3), whereas natural-image shortcuts are non-causal by definition; thus, mitigation must selectively suppress spurious spectral components while preserving causal ones. This balance is difficult without prior knowledge, explaining why augmentation and feature-debiasing can over-penalize. Moreover, spectral SB is more superficial and severe, being directly accessible from raw signal values (P1), while natural-image shortcuts still require some hierarchical abstraction, necessitating more aggressive intervention in our settings. Overall, these approaches are insufficient to overcome our failure. 

That said, the robustification principles behind many of these methods are not inherently irrelevant here. High-frequency overfitting to texture and HSF shortcuts could also arise in IR images; they simply do not manifest because spectral SB collapse is more easily exploitable, prevailing by virtue of their greater simplicity. If this primary issue were mitigated and other failure modes became dominant, such methods might offer greater benefit. At present, however, their impact is limited to marginal gains within the same failure regime.

\subsection{Virtual H\&E incorporation is a strong baseline, but with caveats}\label{s28}
The above limitations highlight the need for caution when applying them to domains where specialized scientific priors interact uniquely with DL's IB. Tailored solutions are therefore necessary. In our context, one effective approach is to incorporate IR-translated virtual H\&E images, which mitigates the failure-inducing properties of raw IR while preserving diagnostic context through RGB morphology. This also enhances compatibility with the mitigation strategies discussed above and with off-the-shelf pathology tools, such as H\&E-based cell segmenters and foundation models \cite{xu2024whole}, while improving interpretability for clinicians. However, this strategy effectively repositions IR imaging as a proxy for H\&E staining,\footnote{Still valuable for reducing labor, tissue damage, staining artifacts, and toxic reagent use \cite{latonen2024virtual}.} underutilizing the rich biochemical information that IR can provide. Notably, we observe that $\sim$6\% of samples were correctly classified only by IR models, corresponding to diagnostically challenging regions unresolved by H\&E morphology alone and highlighting IR’s latent potential. This figure reflects current, suboptimal IR models with limited spectral-spatial learning; improved models could further increase their assistive value.

To leverage the complementary strengths of IR and virtual H\&E, we further investigate strategies that incorporate virtual H\&E priors into the IR model for regularization or fusion. We explore three approaches: (1) {transfer learning}, initializing IR models with weights from virtual H\&E classifiers, using either standard fine-tuning or $\ell_2$-SP \cite{xuhong2018explicit} to constrain weight drift; (2) {multitask learning}, inspired by \cite{liu2023vsgd}, jointly training the IR model for classification and virtual H\&E generation; and (3) \textit{fusion learning}, combining virtual H\&E and IR data via early (image concatenation) or late (final feature concatenation) fusion. Among these, late fusion yields the largest gains, exceeding the performance of standalone virtual (and real) H\&E models. Nonetheless, the improvement remains well below the $\sim$6\% latent potential, indicating room for further enhancement. We test advanced late-fusion strategies, including FiLM (Feature-wise Linear Modulation) \cite{perez2018film}, which modulates features with learned affine transformations, and modality-matching contrastive loss (CL) \cite{yang2024facilitating} to align convergence rates. Neither method surpasses simple concatenation, highlighting the limits of applying these generic strategies to specialized contexts like joint IR–H\&E learning. This underscores the continued need for tailored solutions even in this specific setting.

Despite its promise, there are several caveats to relying on a human-centric RGB counterpart like virtual H\&E, even when fused with the original scientific data. First, it is not always feasible, as translation requires a paired dataset, which may be unavailable, difficult to acquire, or impractical in many scientific contexts. Second, the RGB counterpart may not provide a sufficiently strong IB if it is limited in content, forcing the model to still learn primarily from the scientific modality and potentially reverting to failure. In our study, H\&E provided a good baseline for a relatively simple task like tumor localization. However, for more challenging, fine-grained histopathology tasks, H\&E may be insufficient as it captures only basic morphology. Prior domain knowledge is therefore needed to assess its adequacy, adding further complexity. Together, these limitations indicate that virtual H\&E is not a universal solution, underscoring the need for approaches that can address the failure independently of external, a priori guidance.

\section{Discussion}
\label{s3}
This study was motivated by the hypothesis that generic DL practices may fail in scientific imaging due to misalignment between physiochemical signal priors and the DL’s IBs. This calls into question the common practice of directly adopting such DL vision models as-is, treating them as default starting points for tuning, or using them as so-called ``state-of-the-art'' baselines. We show this through a distinct failure mode in IR-based models, where learning functionally regresses to a quasi-1D spectral analysis rather than robust spectral-spatial 3D reasoning. This arises from an idiosyncratic SB induced by IR spectral characteristics and limits the reliable integration of DL vision with IR imaging. Moreover, because the underlying SB mechanism differs from spurious and shortcut learning in mainstream literature, existing robustification strategies provide limited benefit. Together, these findings highlight the risks of assuming DL methods will transfer reliably from natural image benchmarks to scientific domains. Our results call for more cautious, domain-aware adaptation and underscore the need for tailored DL methodologies in the physical sciences.

DL has achieved remarkable success in approximating human RGB perception. This progress has been driven not only by methodological advances but also by analytical studies that probe how and why DL behaves under specific conditions, including failure modes, thereby informing further innovation. We notice, however, such synergy is largely absent in scientific domains. While many application studies exist, there is a critical lack of work examining how DL fundamentally behaves under non-visible physicochemical signal priors. Addressing this interdisciplinary gap is crucial, and our work takes a step in that direction through a holistic study: we identify a distinct failure mode, trace its origins and mechanisms, characterize its broader consequences, and map the mitigation landscape to clarify the current state of the field and remaining challenges. Such investigations define the present limits of DL in scientific contexts, enabling safer deployment and guiding future methodological development. We advocate for similarly transparent analysis of DL behavior across other scientific imaging modalities—not only failures, but also cases of unexpected success despite IB–prior mismatches, which can be equally informative.

While this work represents an initial step in digital chemical pathology, DL dynamics in this setting remain far from fully characterized. To establish a starting point, we intentionally adopted baseline dataset conditions typical of digital pathology. However, factors such as spectral band selection, modality (e.g., absorbance vs. Raman scattering), dataset size, and task type (e.g., multiclass or segmentation) may substantially influence model behavior and lead to different outcomes. The observed failure hinges on properties P1–P3, which may be weak or absent in other contexts—e.g., when bands are poorly aligned with the pathology objective. A related contemporary study by O’Leary et al. \cite{o2026spatial} demonstrated that spatial context improves DL generalization in IR-based prostate cancer tissue classification, with larger field-of-view inputs and more fine-grained spatial IBs enhancing performance. While this might seem to contradict our findings, the experimental regimes differ. Our work considers an acutely underspecified tile-level classification setting, whereas \cite{o2026spatial} operates at the pixel level with substantially denser supervision, yielding a well-specified problem. In such a regime, we hypothesize that after the initial SB phase, spurious lazy learning is less likely to persist: models can no longer trivially fit the remaining data and are instead driven to learn robust spectral-spatial features. Consequently, our failure mode is not universal, and quantitatively linking it to P1–P3, dataset size, and other factors would be a key step toward preemptive failure identification. Alternative failure modes may also arise: in the absence of low-ED SB collapse, models could overfit to noise or high-spatial-frequency/texture patterns, resembling the high-ED failures commonly observed in natural images. We therefore encourage broader analyses along these axes. Nonetheless, the framework and process-level insights presented here provide a foundation for investigating related DL failure mechanisms beyond IR.

Finally, we acknowledge a potential confound: annotation bias \cite{chen2021understanding,parmar2023don}. All labels were derived from H\&E images, consistent with clinical practice, since pathologists are trained on H\&E slides. Annotations may therefore be implicitly aligned with H\&E-specific visual patterns, potentially favoring H\&E-based models and contributing to the IR-H\&E gap. This does not negate the observed failure mode, as such diagnostic patterns still exist in IR but are systematically underutilized. However, it raises an important open question: how would results differ if supervision were defined directly in IR or other non-H\&E modalities? Addressing this requires careful consideration of gold standards not defined by human assessment or supervision.

\section{Methods}
\label{s4}
\subsection{Dataset} \label{s41}
Our dataset\footnote{This is a subset of a larger dataset from a manuscript currently under preparation. Full cohort and IR acquisition details will be reported there; here we outline only the information necessary for the present study.} comprises 51 frozen prostate tissue sections from 14 patients, sampled across multiple surgical margins at the Mayo Clinic. Multispectral IR data were acquired with a laser-scanning confocal microscope \cite{yeh2023infrared} at mid-IR bands 1081, 1140, 1237, 1275, 1401, 1480, 1547, 1585, 1650, and 1765 cm\textsuperscript{-1}, selected for biological relevance. The same sections were then H\&E-stained, digitized, and registered to the IR data. Slides were downsampled to $\sim$4 µm/pixel to increase field of view, which improved downstream classification in both modalities. Images were tiled into 256×256 pixel\textsuperscript{2} patches, then quality-checked to ensure sufficient tissue presence ($>$50\%) and absence of significant artifacts in both IR and H\&E domains. Tiles failing these criteria were pruned, resulting in $n = 9{,}639$ patches. Each tile was labeled tumorous (31\%) or non-tumorous (69\%) under expert guidance. We used a patient-wise 70/30 train/test split, repeated over $k=15$ random folds to reduce sampling bias.

\subsection{Model Training} \label{s42}
\textit{Baseline classifier.} We use a ResNet50 backbone trained with AdamW \cite{loshchilovdecoupled} and a decaying learning rate initialized at $1\times10^{-4}$, $\beta=(0.9, 0.99)$, and weight decay $1\times10^{-4}$. The batch size is 64, and models are trained for 1,000 iterations with cross-entropy (CE) loss. Inputs are augmented with random horizontal/vertical flips and transpositions. To decouple low-level pixel statistics, all inputs are standardized to approximately $\mathcal{N}(0,1)$.

\textit{Image translation.} For IR-to-H\&E translation and vice versa, we use a Pix2Pix generator \cite{isola2017image} with a multi-scale discriminator \cite{wang2018high}. Models are trained for 15,000 iterations with batch size 16 using adversarial loss combined with an $\ell_1$ reconstruction term and VGG perceptual loss \cite{simonyan2014very}. Optimization uses Adam \cite{kingma2014adam} with an initial learning rate of $2\times10^{-4}$ and $\beta=(0.5, 0.999)$.

\textit{Robust extensions.} Unless noted otherwise, settings follow the baseline. Hyperparameters are tuned empirically or taken from the original papers.

\begin{itemize}[leftmargin=0.5cm] 
\item {Color jitter:} Additive noise $\sim\mathcal{N}(0,0.05)$ is sampled independently per IR band and applied to the Gaussian-standardized channels with probability 0.5. Higher intensities worsened performance.
\item {Cutout} \cite{devries2017improved}: Mask size up to 64×64 pixels, applied with probability 0.25. 
\item {IRMv1} \cite{arjovsky2019invariant} and {V-REx} \cite{krueger2021out}: ``Environments'' are simulated using the color-jitter scheme, mimicking instrument shifts (e.g., laser power variation, baseline drift). To keep memory usage constant, the mini-batch is split into four subsets: three receive independent augmentations to form distinct environments, while one remains unperturbed. Penalty weights are set to 1. 
\item CVaR \cite{levy2020large}: ERM is applied to the top 50\% highest-CE samples in each mini-batch. This is a milder setting than the original paper’s 5–10\%, as these thresholds failed in our context.
\item JTT \cite{liu2021just}: Two-stage training; an initial model is trained, then a second model is trained from scratch with losses of samples misclassified by the first model upweighted by a factor of 2.
\item LfF \cite{nam2020learning}: Two models are trained concurrently: a biased model using Generalized CE \cite{zhang2018generalized} ($q=0.7$), and the main debiased model whose sample-wise loss is weighted by $1 + \frac{CE_B}{CE_B + CE_D}$, where $CE_B$ and $CE_D$ are the sample losses from the biased and debiased models, respectively. Note that the original paper does not include the leading 1; we found it necessary here, as omitting it caused the model to fail.
\item SPARE \cite{yang2024identifying}: Two-stage training: after early stopping at 100 iterations, samples are clustered in the final feature space (cluster number chosen to minimize the silhouette score \cite{rousseeuw1987silhouettes}); training then resumes with importance sampling, where sampling probability is inversely proportional to cluster size.
\item SiFeR \cite{tiwari2023overcoming}: An auxiliary classifier head is attached after the first ResNet block. The loss includes the main CE, the auxiliary CE, and a regularization term—the CE between the auxiliary predictions and a uniform class prior—to suppress shallow spurious features. The regularization term is weighted by 5 and applied every 5 training iterations.
\item ReBias \cite{bahng2020learning}: We adopt a simplified version. An auxiliary two-layer biased model is trained in parallel with standard CE loss, while the main unbiased model is trained with CE plus a Hilbert–Schmidt Independence Criterion (HSIC) penalty between its final features and that of the biased model, encouraging disentanglement. The biased model architecture is: Conv2d($C_{\text{in}}$=10, $C_{\text{out}}$=64, kernel=3) → BatchNorm2d → ReLU → AdaptiveAvgPool2d(1) → Linear head.
\item {ETD} \cite{maini2023can}: We retain a $p_{\text{gen}}=0.4$ fraction of neurons as generalization units and apply dropout with rate $p_{\text{mem}}=0.2$ using fixed, sample-specific neuron masks during training. At test time, only the $p_{\text{gen}}$ neurons are kept active. ETD is applied to the final layer, which is known to be particularly prone to spurious feature encoding \cite{kirichenko2023last,qiu2024complexity}.
\item Focal loss \cite{lin2017focal}: Default parameters $\alpha=0.25$, $\gamma=2.0$.
\item {Mixup} \cite{zhang2018mixup}: Default mixup coefficient sampled from $\sim\text{Beta}(0.2, 0.2)$.
\item {MCD} \cite{gal2016dropout}: Dropout ($p=0.5$) is applied at both training and inference, with predictions marginalized over 5 stochastic forward passes. Dropout is limited to the head for efficiency, a widely used setup in literature.
\item {HLC} \cite{kendall2017uncertainties}: Logits are modeled as Gaussian distributions $(\mu,\sigma)$ with $\text{logit}\sim\mathcal{N}(\mu,\sigma^2)$. Predictions use logits marginalized over 25 samples.
\item M-heads \cite{lee2015m} + ESB \cite{teney2022evading}: Five heads are trained jointly. A diversity regularization term (weight = 1) penalizes similarity between gradients from different heads.
\item ConvNeXt-tiny, \cite{liu2022convnet}, {Swin v2-tiny} \cite{liu2022swin}, {MLP-Mixer-base} \cite{tolstikhin2021mlp}: Batch size is reduced to 32 due to memory constraints.
\item {Virtual H\&E transfer learning}: The first layer is modified for 10-channel input. When used, $\ell_2$-SP regularization \cite{xuhong2018explicit} is applied with coefficient $1\times10^{-4}$.
\item {Virtual H\&E multitask learning}: A classification head is appended to the Pix2Pix decoder for joint IR→H\&E translation and classification. Training follows the standalone translation setup with an additional CE loss.
\item {Virtual H\&E fusion learning}: For early fusion, IR and virtual H\&E images are concatenated into a 13-channel input, and training proceeds normally. For late fusion, each domain is encoded separately by a ResNet50 backbone; the resulting features are fused via simple concatenation or FiLM \cite{perez2018film} and passed to a shared head. Auxiliary heads provide deep supervision \cite{lee2015deeply}. Optionally, a modality-matching contrastive loss \cite{yang2024facilitating} is applied. Batch size is reduced to 32 to accommodate increased memory requirements.
\end{itemize}

\subsection{Cue Manipulation}\label{s43} 
For spatial frequency \cite{solomon1994visual,yin2019fourier,li2023robust,subramanian2024spatial}, we apply a Fast Fourier Transform (FFT) to move images into the frequency domain, mask out a specified frequency range, and then apply the inverse FFT. Following \cite{subramanian2024spatial}, we span seven octave ranges, centered at 1.75, 3.5, 7, 14, 28, 56, and 112 c/i. For HVS cues \cite{geirhos2018generalisation,ge2022contributions,hermann2020origins,mummadidoes}, we perturb images at five severity levels ($\alpha = 1-5$) targeting shape, texture, and color. Specifically, shape is perturbed via random grid-shuffle with grid size $\alpha + 1$, texture is suppressed using Gaussian blur with $\sigma$ sampled from $(0.5\alpha, 0.5(\alpha+1))$, and color is jittered per channel sampled from $[-0.1\alpha, 0.1\alpha]$. Higher $\alpha$ corresponds to stronger perturbation, i.e., greater cue removal. Note, IR and H\&E images are pre-standardized, ensuring comparable relative effect across domains. Each of the 22 manipulations (7 spatial-frequency + 15 HVS) is applied to all test samples, and the resulting change in predictions quantifies the importance of the manipulated cue. While many studies measure accuracy change, we compute $D_{JS}$ over softmax outputs for finer-grained comparisons. HVS sensitivity is reported as the sum over all severity levels. Note, an alternative approach for studying cue bias involves evaluating over cue-conflicting datasets (e.g., a cat shape with elephant texture) \cite{geirhosimagenet}. Constructing such datasets for IR images is non-trivial, so we focus on the more flexible strategy of cue manipulation.

\subsection{Sample Hardness}\label{s44}We adopt several established estimators from the literature to quantify the hardness of individual samples. Arrows ($\uparrow$/$\downarrow$) indicate the direction of increasing hardness.
\begin{itemize}[leftmargin=0.5cm]
\item {Learning speed} \cite{jiang2021characterizing,hacohen2020let} ($\downarrow$): Average accuracy over training; slower convergence implies higher hardness.
\item {Forgetting score} \cite{tonevaempirical,maini2022characterizing} ($\uparrow$): Counts the number of times a sample transitions from correct to incorrect during training.
\item {AUM} \cite{pleiss2020identifying} ($\downarrow$): Average margin between the true logit and largest incorrect logit over training.
\item {EL2N} \cite{paul2021deep} ($\uparrow$): $\ell_2$ norm of the error vector across training; a proxy for gradient magnitude.
\item {Prediction depth} \cite{baldock2021deep} ($\uparrow$): $k$-NN probe ($k=25$) identifies the earliest layer where the sample can be correctly classified. Probed layers: first conv (index 1), final layers of ResNet blocks (indexes 2–5), last conv (index 6).
\item {VoG} \cite{agarwal2022estimating} ($\uparrow$): Variance of gradient magnitudes across training.
\item {Mahalanobis prototypicality} \cite{sorscher2022beyond,lee2018simple} ($\uparrow$): Distance of a sample’s embedding to the nearest Gaussian class centroid from the training set.
\end{itemize}
Metrics based on training dynamics are sampled every 100 iterations (10 checkpoints total). Metrics where lower values indicate higher hardness are inverted, so higher values consistently denote harder samples. All scores are normalized to [0,1] and averaged across metrics to yield a composite hardness value per sample. This composite approach has two advantages: (1) Diversity: Each metric captures a distinct facet of difficulty—AUM reflects confidence margins; EL2N and VoG track gradient dynamics; prediction depth measures representational complexity; prototypicality quantifies embedding density; and learning speed and forgetting scores align with the SB hypothesis, where hard samples are learned late or inconsistently. (2) Reduced bias: Relying on a single metric risks confounding hardness with specific data subgroups. For instance, removing the top 5\% hardest samples in memorization experiments could inadvertently eliminate distribution-level patterns, violating Feldman’s formalism. Combining metrics provides a more nuanced, robust measure of sample difficulty, encompassing mislabeled, ill-posed, atypical, or nonlinear instances.

\subsection{Memorization Experiments}\label{s45}
Following Feldman’s formalism \cite{feldman2020neural}, a sample $(x, y)$ in the $k$-fold training set $\mathcal{D}_k$ is considered memorized if its prediction is correct only when it is included in the training set. The sample-level memorization Score is defined as:
\begin{equation}
mem(x, y; \mathcal{D}_k) = \mathbb{P}(\hat{y}=y \mid \mathcal{D}_k) - \mathbb{P}(\hat{y}=y \mid \mathcal{D}_k^{\setminus {(x,y)}}),
\end{equation}
where $\mathcal{D}_k^{\setminus {(x,y)}}$ denotes the training set with $(x,y)$ removed, $\hat{y}$ is the predicted label, and probabilities are taken over training stochasticity. This score quantifies the excess gain from including the sample, with higher values indicating stronger memorization.
To estimate overall memorization tendency (MT) across domains, we define:
\begin{equation}
MT = \mathbb{E}_{k, (x, y) \in \mathcal{D}_k} \left[ mem(x,y; \mathcal{D}_k) \right].
\end{equation}
Exact computation requires retraining for every leave-one-out sample, which is unfeasible. We adopt subset-based approximations \cite{jiang2021characterizing,lukasiklarger}, removing only the top 5\% hardest training samples per $\mathcal{D}_k$, as these are most likely to be memorized. The remaining 95\%, presumed equally non-memorized or showing minimal cross-domain deviation, are excluded from analysis. Let $H_k \subset \mathcal{D}_k$ denote this hard subset. The MT over the hard subset is thus approximated as:
\begin{equation}
MT_H \approx \mathbb{E}k \Big[\mathbb{P}{(x,y)\in H_k}(\hat{y}=y \mid \mathcal{D}k)- \mathbb{P}{(x,y) \in H_k}(\hat{y}=y \mid \mathcal{D}_k^{\setminus H_k})
\Big].
\end{equation}

\subsection{Dimension Analysis}\label{s46}
\textit{Dimension reduction.} Applied per-pixel for channel-wise compression of IR images. Unsupervised modules (PCA, AE) are fit on a subset of non-background pixel spectra, whereas the supervised module (MLP) is fit on the image's bulk spectra. Separate dimension-reduction modules are fitted per train-test fold to avoid data leakage, and all reduced spectra are further Gaussian-standardized.
\begin{itemize}[leftmargin=0.5cm]
\item {PCA}: 3-component PCA capturing $\sim$91\% cumulative variance.
\item {AE}: Neural net with layer dimensions [10, 8, 5, 3, 5, 8, 10] with BatchNorm1D and ReLU; bottleneck features (3 dimensions) are used as reduced spectra. This provides an improvement over PCA by capturing nonlinearity. Same training setup as baseline, but trained with reconstruction MSE, batch size 512, initial learning rate 1e-2, and weight decay disabled.
\item {MLP}: Neural net with layer dimensions [10, 8, 5, 3, 1] with BatchNorm1D and ReLU; final features (3 dimensions) are used as reduced spectra. Unlike AE, the dimension-reduced features are task-aligned. Trained identically to AE but with CE supervision. 
\end{itemize}

\textit{ED.} Estimated from the ID of the ResNet's final 2048-dimensional feature space. The following ID estimators are considered:
\begin{itemize}[leftmargin=0.5cm]
\item {LPCA} \cite{fukunaga1971algorithm}: Average number of principal components explaining $\sim$95\% variance within each sample's $k \leq 20$ nearest neighbors.
\item {2NN} \cite{facco2017estimating}: Average ratio of distances to 1\textsuperscript{st} and 2\textsuperscript{nd} nearest neighbors; ID inferred from scaling behavior under local uniformity.
\item {MLE} \cite{levina2004maximum}: Generalization of 2NN with $k=6$ nearest neighbors.
\end{itemize}

\subsection{Concordance Score}\label{s47}
We use CAM (Class Activation Mapping) to generate explainable saliency maps $L$ on tumorous patches ($y=1$) to evaluate whether the model attends to any part of the tumor region; otherwise, predictions are not considered causally valid. Let $Y$ denote the pixel-level refined tumor mask. CS is defined as:
\begin{equation}
CS = \mathbb{E}q\left[\mathbb{P}{y=1}(L_q \times Y \neq 0 \mid q)\right],
\end{equation}
where $L_q$ is a binarized version of $L$, thresholded at the top-$q$\% values, with $q\in[90,100]$. Note that IoU or F1 scores are not used, as full tumor coverage is not required—partial identification suffices to classify a patch as tumorous. We adopt Grad-CAM++ \cite{chattopadhay2018grad} for $L$ due to its improved spatial localization and robustness to multiple tumor instances:
\begin{equation}
L = \text{ReLU}\Big(\sum_{k} w_k A_k\Big),
\end{equation}
where $A_k$ are the model’s feature map activations, and $w_k$ are weights derived from higher-order gradients $\partial Y / \partial A_k$. Implementation follows \cite{jacobgilpytorchcam}, using the final layer of the last residual block as $A_k$, as recommended.

\subsection{Uncertainty Quantification}\label{s48}
\textit{ECE.} The Expected Calibration Error (ECE) quantifies how well predicted probabilities reflect true outcome frequencies. It is defined as the expected absolute difference between confidence (predicted probability of the chosen class) and actual accuracy, conditioned on confidence:
\begin{equation}
ECE = \mathbb{E}_{\hat{p}} \left[ \left| \mathbb{P}(y = \hat{y} \mid \hat{p}) - \hat{p} \right| \right],
\end{equation}
where $\hat{y}$ is the prediction, $\hat{p}$ is its confidence, and $y$ is the true label. In practice, this expectation is approximated by discretizing $\hat{p}$ into bins and averaging the empirical differences per bin. We adopt the more robust {SmoothECE} \cite{blasioksmooth}, which smooths the observations using an RBF kernel. Lower ECE values, approaching zero, indicate better-calibrated AU, assuming the samples' EU is low.

\textit{EU.} We focus on DUQ-based EU estimators, requiring no training modifications or test-time ensembles, enabling high-throughput UQ. Recent benchmarks show these methods are highly competitive for EU-related tasks, such as semantic out-of-distribution detection \cite{yang2022openood,oh2024we,mucsanyi2024benchmarking}. Each method below derives uncertainty from different statistics in the model’s final feature or logit space, covering a broad landscape. Many estimators naturally overlap with hardness metrics, as hard samples are associated with high EU. We adopt official implementations, or those from OpenOOD \cite{yang2022openood} or PyOD \cite{chen2025pyod}. Arrows ($\uparrow$/$\downarrow$) indicate the direction of increasing EU.
\begin{itemize}[leftmargin=0.5cm]
\item {ASH} \cite{djurisicextremely} ($\uparrow$): Activation shaping applied before computing Energy \cite{liu2020energy}. We use the ASH-B setup, setting 35\% of activations to a constant and zeroing the remainder.
\item {DML} \cite{zhang2023decoupling} ($\downarrow$): Decouples feature norm, then computes max logit.
\item {RP-GradNorm} \cite{huang2021importance,jiang2023detecting} ($\downarrow$): Gradient norm of the Kullback–Leibler divergence to an empirical class prior estimated from training data.
\item {Mahalanobis} \cite{lee2018simple} ($\uparrow$): Distance to the nearest Gaussian class centroid fitted on training features.
\item {GDA} \cite{mukhoti2023deep} ($\downarrow$): Maximum likelihood under class-conditional Gaussian fitted on training features.
\item {KNN} \cite{sun2022out} ($\uparrow$): Normalized $\ell_2$ distance to the $k=10$ nearest training neighbors in feature space.
\item {Cosine} \cite{techapanurak2020hyperparameter} ($\downarrow$): Feature cosine similarity to the $k=10$ nearest training neighbors.
\item {LUNAR} \cite{goodge2022lunar} ($\uparrow$): Local outlier detection via graph neural network ($k=10$, 200 epochs, otherwise default setting in PYoD) in normalized feature space, treating training samples as inliers.
\item {NNGuide} \cite{park2023nearest} ($\uparrow$): Energy scaled by the above cosine score.
\item {ViM} \cite{wang2022vim} ($\uparrow$): Weighted sum of Energy and the norm of the ``virtual'' logit, defined as the projection onto the subspace orthogonal to the principal subspace of training features ($\sim$95\% variance).
\end{itemize}

\textit{Abstention testing.} As EU is inherently ill-defined, axiomatic experiments are often used to evaluate EU estimators. High-EU samples tend to be miscalibrated, so removing them should improve test-set calibration (ECE) \cite{mucsanyi2024benchmarking}. Let $ECE_q$ denote the ECE after removing the top-$q$\% highest EU samples. An {alignment score (AS)} is defined:
\begin{equation}
AS = \int_0^{100} \frac{ECE_q}{ECE_0} \, dq.
\end{equation}
Normalizing by $ECE_0$ decouples the model’s base ECE, ensuring the metric reflects just EU quality and not AU. Lower AS values indicate better alignment, with 1 corresponding to random rejection. Note, accuracy is sometimes used instead of ECE in the literature; however, rejecting high-AU samples, which tend to be well-calibrated but inaccurate, also improves accuracy, confounding AU and EU evaluation.

\subsection{Other Metrics}\label{s49}
\textit{Cohen's $\kappa$} \cite{geirhos2020beyond}. This metric quantifies the agreement between two models’ predictions, including their consistency in making the same errors, while adjusting for chance agreement. Let $P_o$ denote the observed agreement, i.e., the proportion of samples where both models are either correct or incorrect. Let $P_e$ denote the expected agreement by chance, given by $P_i P_j + (1 - P_i)(1 - P_j)$, where $P_i$ and $P_j$ are the accuracies of the two models being compared. Then:
\begin{equation}
\kappa_{i,j} = \frac{P_o - P_e}{1 - P_e}.
\end{equation}
A $\kappa$ of 1 indicates perfect agreement, 0 indicates chance-level alignment, and values below 0 imply systematic disagreement worse than random.

\textit{CKA} \cite{kornblith2019similarity}. This metric quantifies the similarity between neural feature activations, commonly applied either across models (inter-CKA) or across layers within the same model (intra-CKA). CKA is invariant to orthogonal transformations and isotropic scaling, enabling robust comparison even when feature spaces are misaligned or distributed differently. For a dataset with $N$ samples, let $Z \in \mathbb{R}^{N \times n}$ be the feature matrix, where $n$ is the feature dimension. Define $K = Z\cdot Z^\top$ as the Gram matrix. Given two feature matrices from layers $i$ and $j$, the CKA is computed using HSIC:
\begin{equation}
{CKA}_{i,j} = \frac{{HSIC}(K_i, K_j)}{\sqrt{{HSIC}(K_i, K_i) \cdot {HSIC}(K_j, K_j)}}.
\end{equation}
CKA ranges in $[0,1]$, with higher values indicating greater similarity. For scalability, we adopt the mini-batch approximation from \cite{nguyenwide}.

\textit{Deep SHAP} \cite{lundberg2017unified}. This tool combines DeepLIFT \cite{shrikumar2016not} with Shapley values to approximate input feature attributions in deep networks. We apply it to our trained supervised MLP classifiers in \S\ref{s25} and \S\ref{s46} to estimate the importance of each spectral band. Following the official implementation \cite{lundberg2017unified}, we run Deep SHAP on the full test set using 1,000 stratified training samples as background.

\section{Acknowledgments}
The work reported in this manuscript is supported in part by the National Institutes of Health through grant numbers R01EB009745 and R01CA260830.

\bibliographystyle{ieeetr}
\bibliography{iclr2025_conference}
\end{document}

%% file: math_commands.tex
\usepackage{amsmath,amsfonts,bm}

\def\eqref#1{equation~\ref{#1}}

\def\1{\bm{1}}

\DeclareMathAlphabet{\mathsfit}{\encodingdefault}{\sfdefault}{m}{sl}
\SetMathAlphabet{\mathsfit}{bold}{\encodingdefault}{\sfdefault}{bx}{n}

